# Towards Precision of Probabilistic Bounds Propagation

**Helmut Thöne**
Wilhelm-Schickard-Institut
Universität Tübingen
Sand 13, 7400 Tübingen 1, FRG
thoene@informatik.uni-tuebingen.de

**Ulrich Güntzer**
Wilhelm-Schickard-Institut
Universität Tübingen
Sand 13, 7400 Tübingen 1, FRG
guentzer@informatik.uni-tuebingen.de

**Werner Kießling**
Institut für Informatik
Technische Universität München
Orleansstr. 34, 8 München 80, FRG
wk@informatik.tu-muenchen.de

## Abstract

The DUCK-calculus presented here is a recent approach to cope with probabilistic uncertainty in a sound and efficient way. Uncertain rules with bounds for probabilities and explicit conditional independences can be maintained incrementally. The basic inference mechanism relies on local bounds propagation, implementable by deductive databases with a bottom-up fixpoint evaluation. In situations, where no precise bounds are deducible, it can be combined with simple operations research techniques on a local scope. In particular, we provide new precise analytical bounds for probabilistic entailment.

## 1 Introduction

Numerous approaches for numerical uncertainty have been put forward over the years. Recently we believe to observe a consensus developing within the research community that the days of ad-hoc solutions, like e.g. certainty factors, have passed. Even more formally developed approaches, like e.g. Dempster-Shafer evidence theory, are being considered doubtful, if unsound inferences may be produced at unpredictable times. Instead, fuzzy set methods and various probabilistic systems have gained increased attentions lately ([KSH 91]). We shall focus on the probabilistic methods for uncertain reasoning here. One popular approach are Bayesian networks, with HUGIN as a widely known expert systems shell ([AOJJ 89]). However, Bayesian networks have been criticized on several major aspects: First, unwarranted enforcement of independence assumptions may lead to wrong results. Second, this approach necessitates complete information. This does not only lead to an enormous amount of numerical data to be acquired and managed. More dangerously, if no estimates are available, the principle of indifference (i.e. alternatives are judged to be equally probable) is followed, which may lead to paradox results. And third, always exact probabilities instead of intervals must be supplied. On the strong side, due to above restrictions, efficient local computation procedures producing precise results could be devised ([LaSp 88]).

Alternative approaches, which attempt to overcome the above deficiencies, are represented by the work on the INFERNO-system ([Qui 83]) and by [AnHo 90]. The cautious approach of INFERNO applies a set of inference rules, which only on the basis of explicitly available knowledge draw further sound inferences by *local* bounds propagation. However, their inference process has sometimes been criticized because of too weak bounds ([Pea 88]) or because INFERNO could not succeed to properly control the termination of the inference process ([LiGa 87]). The approach of [AnHo 90] is basicly the same from a user perspective. The difference is in that they propose to translate the problem into linear programming and to apply *global* operations research optimization methods. In this way of course precise bounds can always be inferred. But there is a substantial price to be paid. First there is a loss of explainability of the results to the user. Second, there is an inherent threat of computational intractability as a potentially exponential number of constraints must be optimized. Moreover, the introduction of independence aggravates the problem as it now becomes a non-linear programming task to be solved.

The DUCK approach, presented first in [GKT 91] and [KTG 92], pursues a similar approach at the user interface as INFERNO and [AnHo 90]. However there is a substantial difference in our approach to the optimization of the inference process. Our thesis, which we want to post here, is as follows: We claim that applying global optimization techniques in general not only is inadequate from a knowledge representation standpoint in expert systems, but also is a computational overkill. Therefore we propose an *intelligent combination* of more efficient and comprehensible *local* bounds propagation techniques with *global* optimization tools, where necessarily required. This principle view is taken also by the system proposed by [DPT 90]. In contrast to their work, however, the DUCK approach, described afterwards in more detail, is specifically designed with the objective to map it onto robust and efficient database technology ([KiGu 90],[Ull 89]).

Let us try to add some more evidence to our thesis as claimed above by looking at what's going on in the area of non-monotonic reasoning. This seems reasonable, since non-monotonic phenomena also occur within probabilistic reasoning. Today, several sound and converging theories have evolved, like circumscrip-tion, auto-epistemic logic, modal logic or default reasoning. This is the good news. The bad news is that all these theories are computationally intractable, except for very special cases. The very reason again seems to be the fact that the inference rules used act on a global scope. Thus an intelligent combination of local and global methods, which humans do all the time, seems to be the only way to preserve correctness, but still getting sufficiently precise answers in a reasonable amount of time.

In the sequel we describe the DUCK calculus for uncertain inference in section 2. One detailed example demonstrates that precise results can be obtained by local computation in certain situations. Section 3 is concerned with the problem of precise probabilistic entailment ("rule chaining"). In [DPT 90] a partial solution was already presented. We prove the complete answer here, which interestingly employs some simple operations research techniques on a local scope. Similarly we obtain precise results for probabilistic entailment under independence with intervals. Section 4 summarizes the results obtained with the DUCK approach so far and points out areas of ongoing research.

## 2 The DUCK Calculus for Uncertain Inference

DUCK is an acronym for Deduction of UnCertain Knowledge.

### 2.1 The Probabilistic Calculus

**Definition 2.1 (Conditional probability)**
Let $A, B$ be sets of events and let $AB$ denote the intersection of $A$ and $B$. The conditional probability of $B$ given $A$ is defined as

$$P(B|A) = \frac{P(AB)}{P(A)}, \quad \text{if } P(A) > 0.$$

The equivalent rule-based interpretation is:

$$A \xrightarrow{P(B|A)} B$$

That is $P(B|A)$ among the events in $A$ are also events in $B$. If we have both $A \xrightarrow{x} B$ and $B \xrightarrow{y} A$, we also write $A \xleftarrow{x}{y} B$.

Since precise conditional probabilities are often hard to get or not available, working with intervals makes much more sense in practice:

**Definition 2.2 (Uncertain rule)**
Let $C_1, C_2, \ldots, C_k$ be the set of events. $C_l$ and its complement $\overline{C_l}$, $1 \leq l \leq k$, are called basic events. We consider conjunctive events $A = A_1 \cdots A_n$, $B = B_1 \cdots B_m$, where $n, m \geq 1$ and $A_i, B_j$ are basic events, and $P(A) > 0$. An uncertain rule consists of an upper and a lower bound for a conditional probability:

$$A \xrightarrow{x_1, x_2} B \quad \text{iff} \quad 0 \leq x_1 \leq P(B|A) \leq x_2 \leq 1.$$

If lower and upper bounds coincide we simply write $A \xrightarrow{x} B$.

**Definition 2.3 (Uncertain bidirectional rule)**
Let $A$ and $B$ be events. An uncertain bidirectional rule consists of upper and lower bounds for the conditional probabilities $P(B|A)$ and $P(A|B)$:

$$A \xleftarrow{x_1, x_2}{y_1, y_2} B \quad \text{iff} \quad A \xrightarrow{x_1, x_2} B \text{ and } B \xrightarrow{y_1, y_2} A$$
$$\text{and } (x_2 = 0 \iff y_2 = 0).$$

The requirement $(x_2 = 0 \iff y_2 = 0)$ is reasonable because of $P(B|A) = 0 \iff P(A|B) = 0$. Note that uncertain bidirectional rules with $(x_1 = 0$ and $y_1 > 0)$ or $(x_1 > 0$ and $y_1 = 0)$ are admissible. For instance, in a rule $A \xleftarrow{0, 0.1}{0.8, 1} B$, the value of $P(B|A)$ can be arbitrarily small, but not equal to 0, since $P(A|B) \geq 0.8$ requires $P(B|A) > 0$.

**Definition 2.4 (Inference mechanism)**
Let $\mathcal{R}$ be a set of uncertain rules and conditions consistent with the laws of probability, $A$ and $B$ be conjunctive events.

$$\mathcal{R} \vdash A \xrightarrow{x_1, x_2} B \quad \text{iff} \quad A \xrightarrow{x_1, x_2} B \text{ can be generated,}$$
$$\text{given } \mathcal{R}, \text{ by the following inference}$$
$$\text{rules in a finite number of steps.}$$

**Inference Rules**

($A, B, C$ denote conjunctive events, $F$ denotes a basic event.)

(I1) Chaining (C):

(a) $\{ A \xrightarrow{x_1, x_2} FC, A \xrightarrow{y_1, y_2} \overline{F}C \} \vdash A \xrightarrow{z_1, z_2} C$,
$z_1 = x_1 + y_1, \ z_2 = \min(1, x_2 + y_2)$

(b) $\{ A \xrightarrow{x_1, x_2} BC \} \vdash A \xrightarrow{x_1, 1} C$

(c) $\{ A \xrightarrow{x_1, x_2} BC, C \xrightarrow{1} B \} \vdash A \xrightarrow{x_1, x_2} C$

(d) $\{ A \xrightarrow{x_1, x_2} BC, A \xrightarrow{1} B \} \vdash A \xrightarrow{x_1, x_2} C$

(I2) Sharpening (S):
$\{ A \xrightarrow{x_1, x_2} B, A \xrightarrow{y_1, y_2} B \} \vdash A \xrightarrow{z_1, z_2} B$,
$z_1 = \max(x_1, y_1), \ z_2 = \min(x_2, y_2)$

(I3) Conjunction Left (CL):
$\{ A \xrightarrow{x_1, x_2} B, x_1 > 0, A \xrightarrow{y_1, y_2} BC \} \vdash AB \xrightarrow{z_1, z_2} C$,
$z_1 = \frac{y_1}{x_2}, \ z_2 = \min(1, \frac{y_2}{x_1})$

(I4) Conjunction Right (CR):
$\{ A \xrightarrow{x_1, x_2} B, AB \xrightarrow{y_1, y_2} C \} \vdash A \xrightarrow{z_1, z_2} BC$,
$z_1 = x_1 \cdot y_1, \ z_2 = x_2 \cdot y_2$

(I5) Weak Conjunction Left (WCL):
$\{ A \xleftarrow{x_1, x_2}{} B, x_1 > 0, B \xrightarrow{y_1, y_2} C \} \vdash AB \xrightarrow{z_1, z_2} C$,
$z_1 = \max(0, \frac{x_1 + y_1 - 1}{x_1}), \ z_2 = \min(1, \frac{y_2}{x_1})$

(I6) **Weak Conjunction Right (WCR):**

(a) $\{A \xrightarrow{x_1,x_2} B, A \neq C\} \vdash A \xrightarrow{0,x_2} BC$

(b) $\{A \xrightarrow{x_1,x_2} B, B \xrightarrow{y} C, A \neq C\}$
$\vdash A \xrightarrow{z_1,z_2} BC,$

$z_1 = \begin{cases} 0 & \text{if } y = 0 \\ x_1 & \text{if } y = 1 \end{cases} \quad z_2 = \begin{cases} 0 & \text{if } y = 0 \\ x_2 & \text{if } y = 1 \end{cases}$

(I7) **Negation (N):**

$\{A \xrightarrow{x_1,x_2} F\} \vdash A \xrightarrow{z_1,z_2} \overline{F},$
$z_1 = 1 - x_2, z_2 = 1 - x_1$

(I8) **Conjunction Right with Negation (CRN):**

$\{A \xrightarrow{x_1,x_2} C, A \xrightarrow{y_1,y_2} FC\} \vdash A \xrightarrow{z_1,z_2} \overline{F}C,$
$z_1 = \max(0, x_1 - y_2), z_2 = x_2 - y_1$

(I9) **Weak Conjunction Right with Negation (WCRN):**

$\{A \xleftarrow{u_1,u_2}_{v_1,v_2} F, v_1 > 0,\ F \xleftarrow{x_1,x_2}_{y_1,y_2} C, y_1 > 0, A \neq C\}$
$\vdash A \xrightarrow{z_1,z_2} \overline{F}C,$
$z_1 = 0,\ z_2 = \min(1, (1-y_1) \cdot \frac{u_2 \cdot x_2}{v_1 \cdot y_1})$

(I10) **Annulment (A):**

$\{A \xleftarrow{x_1,x_2}_{0} B, A \xrightarrow{x_1,x_2} B\} \vdash A \xrightarrow{0} B$

Note that above inference rules are *local* in the sense that only some small portions of the uncertain rules in $\mathcal{R}$ or of some already inferenced rules are exploited. By deduction with this calculus the following theorem of sound rule chaining was already derived in [GKT 91].

**Theorem 2.5 (Rule chaining RC)**
Let $A$ and $C$ be conjunctive events, $B$ a basic event.
$\mathcal{R} = \{A \xleftarrow{u_1,u_2}_{v_1,v_2} B, B \xleftarrow{x_1,x_2}_{y_1,y_2} C\}.$ Then $\mathcal{R} \vdash A \xrightarrow{z_1,z_2} C,$

$z_1 = \begin{cases} \frac{u_1}{v_1} \cdot \max(0, v_1 + x_1 - 1) & \text{if } v_1 > 0 \\ u_1 & \text{if } v_1 = 0,\ x_1 = 1 \\ 0 & \text{otherwise.} \end{cases}$

$z_2 = \begin{cases} \min(1, u_2 + \tau \cdot (1 - y_1), 1 - u_1 + \tau \cdot y_1, \tau) \\ \quad \text{with } \tau = \frac{u_2 x_2}{v_1 y_1} \text{ if } v_1 > 0,\ y_1 > 0 \\ \min(1, 1 - u_1 + \frac{u_2 x_2}{v_1}) & \text{if } v_1 > 0,\ y_1 = 0 \\ 1 - u_1 & \text{if } v_1 = 0,\ x_2 = 0 \\ 1 & \text{otherwise.} \end{cases}$

### 2.2 Conditional Independence

Most other calculi have their difficulties when it comes to deal with conditional independence information: either they cannot handle it (e.g. the SIMUNC system of [Rim 90]) or they suppose it with necessity ([AOJJ 89]). The DUCK calculus allows us to enter such information explicitly into the system and correlates it with the remaining uncertain rules by the following natural extension.

**Definition 2.6 (Conditional independence)**
Let $A, B$ and $C$ be events. $C$ is independent of $A$ under condition $B$, denoted $I(A, B, C)$, iff $P(C|BA) = P(C|B)$.

The requirement $P(C|BA) = P(C|B)$ is equivalent to $P(AB) > 0$ and $P(AC|B) = P(A|B) \cdot P(C|B)$.

(I11) **Invariance (I):**

(a) $\{B \xrightarrow{x_1,x_2} C, I(A,B,C)\} \vdash AB \xrightarrow{x_1,x_2} C$

(b) $\{AB \xrightarrow{x_1,x_2} C, I(A,B,C)\} \vdash B \xrightarrow{x_1,x_2} C$

(I12) **Symmetry (SYM):** [1]

$\{I(A,B,C),\ B \xleftarrow{x_1,x_2}_{y_1,y_2} C,\ x_1 > 0 \text{ or } y_1 > 0\}$
$\vdash I(C, B, A)$

Conditional independence information often makes the problem more constrained and can lead to sharper intervals. [2]

**Theorem 2.7 (Rule chaining under independence)**
Let $A$ and $C$ be conjunctive events, $B$ a basic event,
$\mathcal{R} = \{A \xrightarrow{u} B, B \xrightarrow{x} C, \overline{B} \xrightarrow{y} C, I(A,B,C), I(A,\overline{B},C)\}$
and let $w = u \cdot x + (1-u) \cdot y$. Then

(RCI1) $\quad \mathcal{R} \vdash A \xrightarrow{w} C$

(RCI2) $\quad \mathcal{R} \cup \{w > 0\} \vdash AC \xrightarrow{z} B,\ z = \frac{u \cdot x}{w}$

The proof is again by *local* deduction with the DUCK calculus ([KTG 92]). Evidently the bounds for $w$ and $z$ are precise, because we have derived a 1-point interval.

**Example (Metastatic cancer)** (cf. [Spie 86])
Metastatic cancer is a possible cause of a brain tumor and is also an explanation for increased total serum calcium. In turn, either of these could explain a patient falling into a coma. Severe headache is also possibly associated with a brain tumor. Figure 1 shows the diagram representing these causal influences among others.

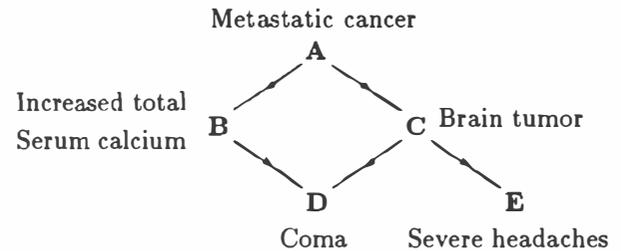

Figure 1: Causal diagram for metastatic cancer

---

[1] In [Pea 88] further inference rules are presented for deducing new independences from a given set of independences.

[2] The former rule named (IND) in [KTG 92] was found to be redundant in the meanwhile.

318   Thöne, Güntzer, and Kießling

These influences are expressed in terms of conditional probabilities and additional independences:

$$\mathcal{R} = \{A \xrightarrow{0.8} B, A \xrightarrow{0.2} C, BC \xrightarrow{0.8} D, \overline{B}C \xrightarrow{0.8} D,$$
$$B\overline{C} \xrightarrow{0.8} D, \overline{B}\,\overline{C} \xrightarrow{0.05} D, C \xrightarrow{0.8} E, \overline{C} \xrightarrow{0.6} E,$$
$$I(A,BC,D), I(A,\overline{B}C,D), I(A,B\overline{C},D),$$
$$I(A,\overline{B}\,\overline{C},D), I(A,C,E), I(A,\overline{C},E),$$
$$I(B,A,C), I(\overline{B},A,C)\}$$

With the rule chaining theorem 2.7 we can directly deduce:

$$\{A \xrightarrow{0.2} C, C \xrightarrow{0.8} E, \overline{C} \xrightarrow{0.6} E, I(A,C,E), I(A,\overline{C},E)\}$$
$$\vdash_{\text{RCII}} A \xrightarrow{0.64} E$$

Even for the more complicated conclusion $A \xrightarrow{?} D$, applying inference rules I, N, CR, C properly we get the sharp result $A \xrightarrow{0.68} D$:

$$\mathcal{R} \vdash_{I,N} \mathcal{U}_1 = \{ABC \xrightarrow{0.8} D, A\overline{B}C \xrightarrow{0.8} D, AB\overline{C} \xrightarrow{0.8} D,$$
$$A\overline{B}\,\overline{C} \xrightarrow{0.05} D, AB \xrightarrow{0.2} C, A\overline{B} \xrightarrow{0.2} C,$$
$$A \xrightarrow{0.2} \overline{B}, A \xrightarrow{0.8} \overline{C}\}$$

$$\mathcal{U}_1 \vdash_{\text{CR},N} \mathcal{U}_2 = \{AB \xrightarrow{0.16} CD, A\overline{B} \xrightarrow{0.16} CD,$$
$$AB \xrightarrow{0.8} \overline{C}, A\overline{B} \xrightarrow{0.8} \overline{C}\}$$

$$\mathcal{U}_1 \cup \mathcal{U}_2 \vdash_{\text{CR}} \mathcal{U}_3 = \{AB \xrightarrow{0.64} \overline{C}D, A\overline{B} \xrightarrow{0.04} \overline{C}D\}$$

$$\mathcal{U}_2 \cup \mathcal{U}_3 \vdash_{C} \mathcal{U}_4 = \{AB \xrightarrow{0.8} D, A\overline{B} \xrightarrow{0.2} D\}$$

$$\mathcal{R} \cup \mathcal{U}_1 \cup \mathcal{U}_4 \vdash_{\text{CR}} \mathcal{U}_5 = \{A \xrightarrow{0.64} BD, A \xrightarrow{0.04} \overline{B}D\}$$

$$\mathcal{U}_5 \vdash_{C} \mathcal{U}_6 = \{A \xrightarrow{0.68} D\}$$

To contrast this sort of local deduction with the global operations research optimization approach, the reader is referred to [AnHo 90] and [KTG 91].

## 3   Precise Bounds for Probabilistic Entailment

So far we have been busy in emphasizing only precise bounds gained by local DUCK deduction. However, a closer look reveals situations, where no precise bounds can be obtained in this way.

Let us refer back to the rule chaining theorem RC, which we deliberately have not claimed to be precise under all circumstances. This would not have been true in general. In [DuPr 88], [DPT 90] other rule chaining theorems (called "fuzzy syllogisms" there) are stated, which are also not precise (or "locally complete" in the terminology of [DPT 90]). In the following we shall correct this and moreover we can provide precise bounds for extreme cases not described in the literature so far. The latter is very important from the point of view of building *robust expert system software*.

### 3.1   The Precise Rule Chaining Theorem

Accuracy of bounds in uncertainty reasoning is of course a relative notion depending on the underlaying uncertainty model. The term "precise" we employ here is meant to apply to the conditional probability model.

**Definition 3.1 (Precision)**
*Let $\mathcal{R}$ be a set of uncertain rules and independences, $A$ and $B$ be events.*
$\mathcal{R} \models_{\text{precise}} A \xrightarrow{z_1, z_2} B$ *iff $z_1$ is the greatest lower bound and $z_2$ is the least upper bound of all bounds for $P(B|A)$ that follow from $\mathcal{R}$ and the laws of probability.*

**Theorem 3.2 (Precise rule chaining (PRC))**
*Let $A, B, C$ be conjunctive events.*
$$\{A \xleftrightarrow[v_1,v_2]{u_1,u_2} B, B \xleftrightarrow[y_1,y_2]{x_1,x_2} C\} \models_{\text{precise}} A \xrightarrow{z_1,z_2} C \quad \text{for}$$

$$z_1 = \begin{cases} \max(0, u_1 \cdot (1 - \frac{1}{v_1} \cdot (1-x_1))) & \text{if } v_1 > 0 & [1] \\ u_1 & \text{if } v_1 = 0, x_1 = 1 & [2] \\ 0 & \text{otherwise.} & [3] \end{cases}$$

$$z_2 = \begin{cases} \min(1, \frac{u_2 x_2}{v_1 y_1}, u_2(1-\frac{x_2}{v_1}(1-\frac{1}{y_1})), 1-u_1(1-\frac{x_2}{v_1}), \\ \quad \frac{x_2}{y_1(v_1-x_2)+x_2}) & \text{if } v_1 > 0, y_1 > 0 & [4] \\ \min(1, 1-u_1(1-\frac{x_2}{v_1})) & \text{if } v_1 > 0, y_1 = 0 & [5] \\ 1 - u_1 & \text{if } v_1 = 0, x_2 = 0 & [6] \\ u_2 & \text{if } v_1 = 0, y_1 = 1 & [7] \\ 1 & \text{otherwise.} & [8] \end{cases}$$

In comparison with [DuPr 88], [DPT 90], above theorem additionally contains the extreme cases [2], [3], [6], [7] and [8]. Moreover case [4] improves the bound given in [DPT 90] by the last stated term.

Compared to theorem RC, the cases [4], [5] and [7] improve the bounds given in (2.5).

In order to get a compact proof of PRC, we start over from results given already by [DPT 90]: For *precise* conditional probabilities $u = P(B|A)$, $v = P(A|B)$, $x = P(C|B)$ and $y = P(B|C)$ the following holds:

Given uncertain bidirectional rules $A \xleftrightarrow[v]{u} B$ and $B \xleftrightarrow[y]{x} C$, there exist events $A_1, B_1, C_1$ and $A_2, B_2, C_2$ such that

$$P(C_1|A_1) = \max(0, u \cdot (1 - \frac{1}{v} \cdot (1-x))) \quad \text{if } v > 0 \quad (*)$$

$$P(C_2|A_2) = \min(1, \frac{ux}{vy}, u(1-\frac{x}{v}(1-\frac{1}{y})), 1-u(1-\frac{x}{v}))$$
$$\text{if } v > 0, y > 0 \quad (**)$$

where equation (*) provides the greatest lower bound and equation (**) provides the least upper bound.

Based on this result we make the correct transition to intervals, i.e. for all cases ([1]-[8]), we consider $u \in [u_1, u_2]$, $v \in [v_1, v_2]$, $x \in [x_1, x_2]$ and $y \in [y_1, y_2]$ with the constraints $(u > 0 \iff v > 0)$ and $(x > 0 \iff y > 0)$. This transition was partially erroneous in [DPT 90].

**Proof: $z_1$ is the greatest lower bound**

Case [1]: Minimizing $u, v, x$ will minimize the formula in (*) as stated already in [DuPr 88], verifying [1]. More precisely we can observe: If $u_1 > 0$ and $x_1 > 0$, then $z_1$ is a *minimum*. On the other hand, if $u_1 = 0$ or $x_1 = 0, z_1$ can be either a *minimum* or an *infimum*. E.g. $z_1$ is an *infimum* in the following case: $u_1 = 0, x_1 > 0$ and $v_1 > 1 - x_1$. Since $v_1 > 1 - x_1 \geq 0$, we get $\frac{1}{v_1}(1-x_1) < 1$ and additionally $u > 0$ because of $v_1 > 0$. Therefore

$$u \cdot (1 - \tfrac{1}{v_1}(1 - x_1)) > 0.$$

Since $u > u_1 = 0$, we can choose $u$ arbitrarily small and we get:

$$\lim_{u \to 0} u \cdot (1 - \tfrac{1}{v_1}(1 - x_1)) = 0 = z_1.$$

That is, $z_1$ is an *infimum* which can be approached arbitrarily close, but never reached exactly.

Case [2]: Since $x_1 = 1$, we choose $B_1, C_1$ with $B_1 C_1 = B_1$, i.e. $B_1 \subseteq C_1$ in any case.
(i) $u_1 = 0$: Because of $v_1 = 0$ we can choose $A_1$ being disjoint with $C_1$, yielding $z_1 = P(C_1|A_1) = 0 = u_1$.
(ii) $u_1 > 0$: For consistency reasons we get $v > 0$. This allows us to apply [1] with $x = x_1 = 1$ and $u = u_1$ resulting in $z_1 = u_1$.

Case [3]: This case covers $v_1 = 0$ and $x_1 < 1$.
(i) $u_1 = 0$: We choose $A_1, B_1, C_1$ and $x \geq x_1, y \geq y_1$ such that $P(C_1|B_1) = x$, $P(B_1|C_1) = y$ and $A_1$ is disjoint from $B_1$ and $C_1$, i.e. $z_1 = P(C_1|A_1) = 0$:

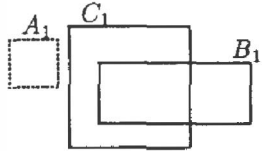

(ii) $u_1 > 0$: Consistency again implies $v > 0$, so [1] applies. We can choose $v$ very small such that $v < 1 - x$, yielding $z_1 = P(C_1|A_1) = 0$.

This completes the proof for $z_1$. ∎

**Proof: $z_2$ is the least upper bound**

Case [4]: We observe in (**) that maximizing $x$ will maximize each term and minimizing $v$ and $y$ will maximize each term in (**). However, considering $u$ reveals a different behavior. So we are left to optimize the following problem:

$$\max_u (\min(1, \tfrac{u x_2}{v_1 y_1}, u(1 - \tfrac{x_2}{v_1}(1 - \tfrac{1}{y_1})), 1 - u(1 - \tfrac{x_2}{v_1})))$$

The second and third term are increasing linearly in $u$, while the fourth term is decreasing linearly in $u$ if $x_2 < v_1$ and is greater or equal to 1 otherwise.
For $x_2 < v_1$, we have:

$$u(1 - \tfrac{x_2}{v_1}(1 - \tfrac{1}{y_1})) = u(1 - \tfrac{x_2}{v_1}) + \tfrac{u x_2}{v_1 y_1} \geq \tfrac{u x_2}{v_1 y_1}$$

Thus we consider only the second term and fourth term under the constraints $v_1 > 0$, $y_1 > 0$ and $x_2 < v_1$. Both terms yield the same result at $u_o$:

$$\tfrac{u_o x_2}{v_1 y_1} = 1 - u_o(1 - \tfrac{x_2}{v_1}) \iff u_o = \tfrac{v_1 y_1}{y_1(v_1 - x_2) + x_2}$$

If $u_o \epsilon [u_1, u_2]$ we get the solution:

$$\tfrac{u_o x_2}{v_1 y_1} = \tfrac{x_2}{y_1(v_1 - x_2) + x_2}$$

The denominator $y_1(v_1 - x_2) + x_2 = v_1 y_1 + x_2(1 - y_1)$ is positive for $v_1 > 0$, $y_1 > 0$ and the value of this term is greater or equal to the values of the other terms when $u_o \notin [u_1, u_2]$ or when we have the constraints $v_1 > 0$, $y_1 > 0$ and $x_2 \geq v_1$. Therefore we can add the new term to the other min-terms.

Case [5]:
(i) $x_2 = 0$: We can choose $A_2, B_2, C_2$ and $u \geq u_1, v \geq v_1$ such that $P(B_2|A_2) = u$, $P(A_2|B_2) = v$, $B_2$ is disjoint from $C_2$ and $A_2 \overline{B_2} = A_2 C_2$ because $C_2$ can be arbitrarily big:

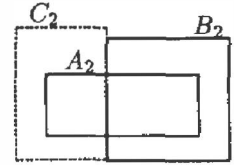

Due to $A \xrightarrow{1-u_2, 1-u_1} \overline{B}$, we get $z_2 = 1 - u_1$, which is a special case of [5] for $x_2 = 0$.

(ii) $x_2 > 0$: This implies $y_2 > 0$ and $y > 0$ can be assumed (case $y = 0$ is discussed in [5i]). Because now $v_1$ and $y$ are positive, we can apply [4] with

$$z_2 = \lim_{y_1 \to 0} [4]$$
$$= \min(1, +\infty, +\infty, 1 - u_1(1 - \tfrac{x_2}{v_1}), 1)$$
$$= \min(1, 1 - u_1(1 - \tfrac{x_2}{v_1})).$$

To be more accurate: if $u_1 = 0$ and $x_2 < v_1$ then $z_2$ is a *supremum* for $P(C_2|A_2)$ and otherwise a *maximum* for $P(C_2|A_2)$.

Case [6]: Due to $x_2 = 0$ we get $y_2 = 0 = y_1$, i.e. $B_2$ and $C_2$ are disjoint.
(i) $u_1 = 0$: Since $C_2$ can be arbitrarily large and $v_1 = 0$, we can choose $A_2$ being disjoint from $B_2$ as follows:

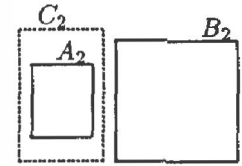

This implies $z_2 = P(C_2|A_2) = 1 = 1 - u_1$.

(ii) $u_1 > 0$: Due to $A \xrightarrow{1-u_2, 1-u_1} \overline{B}$ we get $z_2 = 1 - u_1$, like in case [5i].

Case [7]:
(i) $u_2 = 0$: We must choose $A_2, B_2, C_2$ such that $A_2$ is disjoint from $B_2$ and $B_2 \supseteq C_2$:

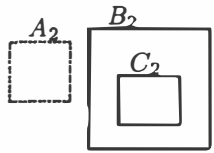

Thus $z_2 = P(C_2|A_2) = 0 = u_2$.

(ii) $u_2 > 0$: Now $u > 0$ can be assumed (case $u = 0$ is discussed in [7i]), hence $v > 0$. Since $y_1 = 1 > 0$ and $v > 0$ we can apply [4] as follows:

$P(C_2|A_2) = \min(1, \frac{u_2 x_2}{v_1}, u_2, 1 - u_1(1 - \frac{x_2}{v_1}), \frac{x_2}{v_1})$

Thus

$$\begin{aligned} z_2 &= \lim_{v_1 \to 0} P(C_2|A_2) \\ &= \min(1, +\infty, u_2, +\infty, +\infty) \\ &= u_2. \end{aligned}$$

Case [8]: The remaining case covers $v_1 = 0$, $x_2 > 0$, $y_1 < 1$. Consistency implies $y_2 > 0$.
(i) $u_2 = 0$: We assume $y > 0$ and choose $A_2, B_2, C_2$ such that $P(C_2|B_2) = x_2$, $P(B_2|C_2) = y$, $A_2$ is disjoint from $B_2$ and $A_2 \overline{B_2} = A_2 C_2$:

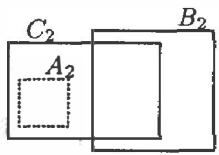

This is due to the fact that $A_2$ can be arbitrarily small and $\overline{B_2} C_2$ is not empty because of $P(\overline{B_2}|C_2) \geq 1 - y_1 > 0$. Hence $z_2 = 1$.

(ii) $u_2 > 0$: Thus $v_2 > 0$ is implied. We can assume $v > 0$ and $y > 0$ and apply [4] with

$z_2 = \lim_{v_1 \to 0} [4] = \min(1, +\infty, +\infty, +\infty, \frac{1}{1-y_1}) = 1.$

Note that it suffices to prove that $z_2 = 1$ for the assumption $v > 0$ and $y > 0$.

This completes the proof of $z_2$. ∎

**Example (Precise Rule Chaining)**

$\{A \xrightarrow[.8,.8]{.2,.8} B, B \xrightarrow[.2]{.2} C\} \models_{\overline{precise}} A \xrightarrow{0,.625} C$

It is interesting to observe how simple operations research techniques enter the scene here *at a local scope*:

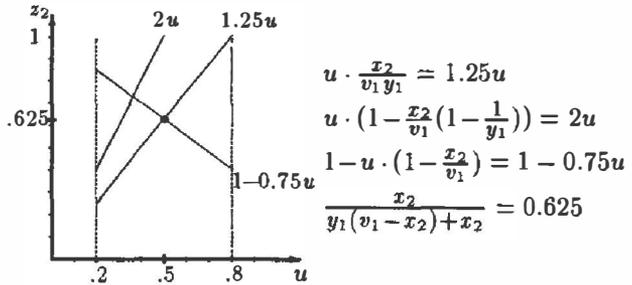

Further examples of precise rule chaining are listed in the figure 2 below.

| | | PRC | [DPT 90] |
|---|---|---|---|
| [4]: | $A \xleftrightarrow[1,1]{0,1} B \xleftrightarrow[.01]{.01} C$ | $A \xrightarrow{0,.50} C$ | $A \xrightarrow{0,1} C$ |
| [4]: | $A \xleftrightarrow[.9,.9]{.1,.9} B \xleftrightarrow[.1]{.1} C$ | $A \xrightarrow{0,.56} C$ | $A \xrightarrow{0,.91} C$ |
| [4]: | $A \xleftrightarrow[.8,1]{.1,1} B \xleftrightarrow[.3,1]{.4,.4} C$ | $A \xrightarrow{.03,.77} C$ | $A \xrightarrow{.03,.95} C$ |
| [4]: | $A \xleftrightarrow[1,1]{.6,1} B \xleftrightarrow[.8]{.8} C$ | $A \xrightarrow{.48,.83} C$ | $A \xrightarrow{.48,.88} C$ |
| [4]: | $A \xleftrightarrow[1,1]{.8,1} B \xleftrightarrow[.95]{.95} C$ | $A \xrightarrow{.76,.95} C$ | $A \xrightarrow{.76,.96} C$ |
| [2]: | $A \xleftrightarrow[0,1]{.9,1} B \xleftrightarrow[0,1]{1,1} C$ | $A \xrightarrow{.9,1} C$ | - |
| [6]: | $A \xleftrightarrow[0,1]{.9,1} B \xleftrightarrow[0,0]{0,0} C$ | $A \xrightarrow{0,.1} C$ | - |
| [7]: | $A \xleftrightarrow[0,1]{0,.2} B \xleftrightarrow[1,1]{0,1} C$ | $A \xrightarrow{0,.2} C$ | - |

Fig.2: Comparison of PRC and results in [DPT 90].

The correctness of these results have been verified by running SIMUNC ([Rim 90]), working with global operations research methods, on above sample data.
PRC can also support the detection of inconsistencies in the case of rule chaining in cycles ([Hei 91],[GKT 91]).

### 3.2 Precise Rule Chaining Under Independence

The previously given theorem (2.7) for rule chaining under independence was a special case in so far as only point probabilities were considered. If this restriction is relaxed, then again by doing only local DUCK deduction we cannot guarantee precise answers under all circumstances. [ADP 91] analyze a related problem and provide some preliminary results. Subsequently we will focus on a special variant of rule chaining under independence.

**Theorem 3.3 (Precise rule chaining under independence with intervals)**
Let $A, B$ and $C$ be events and $\mathcal{R}$ be

$\{A \xrightarrow{u_1, u_2} B, B \xrightarrow{x_1, x_2} C, \overline{B} \xrightarrow{y_1, y_2} C, I(A,B,C), I(A,\overline{B},C)\}$.

*Then*

(a) $\quad \mathcal{R} \models_{\overline{precise}} A \xrightarrow{z_1, z_2} C \quad \text{for}$

$$z_1 = \begin{cases} u_1 \cdot x_1 + (1 - u_1) \cdot y_1 & \text{if } x_1 > y_1 \\ u_2 \cdot x_1 + (1 - u_2) \cdot y_1 & \text{otherwise.} \end{cases}$$

$$z_2 = \begin{cases} u_2 \cdot x_2 + (1 - u_2) \cdot y_2 & \text{if } x_2 > y_2 \\ u_1 \cdot x_2 + (1 - u_1) \cdot y_2 & \text{otherwise.} \end{cases}$$

(b) $\quad \mathcal{R} \cup (x_1 > 0 \text{ or } y_1 > 0) \models_{\overline{precise}} AC \xrightarrow{z_1, z_2} B \quad \text{for}$

$$z_1 = \begin{cases} 1 & \text{if } u_1 = 0, y_2 = 0 \\ \frac{u_1 \cdot x_1}{u_1 \cdot x_1 + (1 - u_1) \cdot y_2} & \text{otherwise} \end{cases}$$

$$z_2 = \begin{cases} 0 & \text{if } u_2 = 1, x_2 = 0 \\ \frac{u_2 \cdot x_2}{u_2 \cdot x_2 + (1 - u_2) \cdot y_1} & \text{otherwise} \end{cases}$$

**Proof:**

(a) For point probabilities $u = P(B|A)$, $x = P(C|B)$ and $y = P(C|\overline{B})$, theorem 2.7 implies $P(C|A) = u \cdot x + (1-u) \cdot y = u \cdot (x-y) + y$. Then making the transition to intervals, i.e. $u \in [u_1, u_2]$, $x \in [x_1, x_2]$ and $y \in [y_1, y_2]$, the following optimization problems arise:

$$z_1 = \min_{u, x, y} \quad (u \cdot (x - y) + y)$$

$$z_2 = \max_{u, x, y} \quad (u \cdot (x - y) + y)$$

As both expressions are linear in $x, y$ and $u$, we immediately get to the stated result.

(b) The conditional independences $I(A, B, C)$ and $I(A, \overline{B}, C)$ imply $P(AB) > 0$ and $P(A\overline{B}) > 0$. And we can assume $u_2 > 0$ and $u_1 < 1$, because we always require our rule base $\mathcal{R}$ to be consistent within the laws of probability.

Again we use the result from theorem 2.7. The optimization problems are now:

$$z_1 = \min_{u, x, y} \quad \left(\frac{u \cdot x}{u \cdot x + (1-u) \cdot y}\right)$$

$$z_2 = \max_{u, x, y} \quad \left(\frac{u \cdot x}{u \cdot x + (1-u) \cdot y}\right)$$

Since $P(AB) > 0$, we have $u = P(B|A) > 0$ and this allows us to conclude

$$\frac{u \cdot x}{u \cdot x + (1-u) \cdot y} = \frac{x}{x + \frac{(1-u)}{u} \cdot y}$$

which is increasing monotonically in $u, x$ and decreasing monotonically in $y$. This proves the two otherwise-cases for $z_1$ and $z_2$.

In the special case $u_1 = 0$ and $y_2 = 0$ we get:

$$z_1 = \frac{u \cdot x_1}{u \cdot x_1 + 0} = 1,$$

whereas in the case $u_2 = 1$ and $x_2 = 0$ we get:

$$z_2 = \frac{0}{0 + (1 - u) \cdot y_1} = 0.$$

In all other cases, the denominators are positive. This completes the proof. ∎

**Example (Precise rule chaining under independence)**

Consider a causal chain

$$A \rightarrow B \rightarrow C \rightarrow D$$

with $A \xrightarrow{.8, 1} B$, $B \xrightarrow{.7, .8} C$, $\overline{B} \xrightarrow{.2, .3} C$, $C \xrightarrow{.4} D$ and $\overline{C} \xrightarrow{.8, .9} D$. First we calculate the belief of event C incorporating the causal support from $A$, whereas then we update the belief in C by including the diagnostic support from $D$:

$$\{A \xrightarrow{.8, 1} B, B \xrightarrow{.7, .8} C, \overline{B} \xrightarrow{.2, .3} C, I(A, B, C), I(A, \overline{B}, C)\}$$

$$\models_{\overline{precise}} A \xrightarrow{.6, .8} C$$

$$\{A \xrightarrow{.6, .8} C, C \xrightarrow{.4} D, \overline{C} \xrightarrow{.8, .9} D, I(A, C, D), I(A, \overline{C}, D)\}$$

$$\models_{\overline{precise}} AD \xrightarrow{.4, .67} C$$

This example demonstrates that accumulated intervals do not necessarily diffuse towards $[0, 1]$ in causal chains.

## 4  Summary and Outlook

We have presented the DUCK approach to probabilistic reasoning. A small set of sound, local inference rules - including explicit conditional independence, comparative probabilities and absolute probabilities [KTG 92] - often achieves to deduce precise bounds for probabilistic queries, including conjunction or negation. It is our thesis that only where precision cannot be gained by locality more global algorithms based on operations research should be employed. In this way we stated precise analytical results for two forms of rule chaining. The DUCK approach is tailored to relational and deductive database technology with a bottom-up fixpoint evaluation. Factual knowledge residing in commercial databases can be combined with uncertain rules, based on ideas of maximal context and detachment ([TGK 91]).

Our long-term objective with the DUCK approach is to bring uncertainty reasoning into a practical database environment for real-life applications. There are, however, several open problems of substantial difficulty to be solved until this ambitious goal can be realized. A static compile-time analysis to detect, when local deductions are not sufficient and what more global algorithms to use then, is a major next challenge.

### Acknowledgements

We would like to thank the anonymous referees for helpful comments.